%% file: main.tex
\documentclass{article} % For LaTeX2e
\usepackage{iclr2026_conference,times}

\input{math_commands.tex}

\usepackage{hyperref}
\usepackage{url}
\usepackage{booktabs}
\usepackage{booktabs}

\title{A Single Image and Multimodality Is All You Need for Novel View Synthesis}

\author{Amirhosein Javadi, Chi-Shiang Gau, Konstantinos D. Polyzos, Tara Javidi \\
Department of Electrical and Computer Engineering\\
University of California San Diego\\
\texttt{\{amjavadi,cgau,kpolyzos,tjavidi\}@ucsd.edu} \\
}

\usepackage{graphicx}
\usepackage{booktabs}
\usepackage{caption}

\iclrfinalcopy % Uncomment for camera-ready version, but NOT for submission.
\begin{document}

\maketitle

\input{Sec/1_abstract}
\input{Sec/2_Intro}
\input{Sec/4_Method}
\input{Sec/5_Experiments}

\section{Conclusions}
We introduced a multimodal framework for single-image novel-view synthesis as an efficient and reliable alternative to vision-only diffusion-based counterpart. In addition to the visual information from the single view, we proposed a depth map reconstruction approach by  
modeling sparse radar measurements via a computationally efficient principled localized Gaussian Process framework producing dense depth maps with spatially varying uncertainty. The  estimated depth integrates seamlessly with existing diffusion-based rendering pipelines, improving geometric consistency and alignment across viewpoints. Experiments on diverse scenes from a real-world multimodal autonomous-driving database demonstrated that our approach resulted in significant improvements in downstream video generation quality compared to image-only baselines; hence justifying the claim that \textbf{a single image and multimodality is all you need for efficient 3D scene perception.} In our future research agenda, we aim to explore the benefits of the proposed depth and uncertainty representation for broader multimodal 3D perception tasks, including mapping, planning, and sensor fusion. 
%Overall, this work underscores the importance of incorporating reliable non-visual modalities into generative 3D pipelines and supports a central takeaway: \textbf{a single image and multimodality is all you need for efficient 3D scene perception.}

\section*{Acknowledgements}
This work was supported by the Eric and Wendy Schmidt AI for Science, the NSF TILOS AI Institute, the UCSD Centers for Machine intelligence, computing, and security (MICS) and Wireless Communications (CWC), the computational resources from Amazon Web Services (AWS), the ONR Award N00014-22-1-2363 and the NSF grant 2148313, with the latter being supported in part by funds from federal agency and industry partners as specified in the Resilient \& Intelligent NextG Systems (RINGS) program.

\bibliography{iclr2026_conference}
\bibliographystyle{iclr2026_conference}

% \appendix
% \section{Appendix}

\end{document}

%% file: math_commands.tex
\usepackage{amsmath,amsfonts,bm}

\def\eqref#1{equation~\ref{#1}}
\def\1{\bm{1}}

\DeclareMathAlphabet{\mathsfit}{\encodingdefault}{\sfdefault}{m}{sl}
\SetMathAlphabet{\mathsfit}{bold}{\encodingdefault}{\sfdefault}{bx}{n}

%% file: Sec/1_abstract.tex
\begin{abstract}
Diffusion-based approaches have recently demonstrated strong performance for single-image novel view synthesis by conditioning generative models on geometry inferred from monocular depth estimation. However, in practice, the quality and consistency of the synthesized views are fundamentally limited by the reliability of the underlying depth estimates, which are often fragile under low texture, adverse weather, and occlusion-heavy real-world conditions. In this work, we show that incorporating sparse multimodal range measurements provides a simple yet effective way to overcome these limitations. We introduce a multimodal depth reconstruction framework that leverages extremely sparse range sensing data, such as automotive radar or LiDAR, to produce dense depth maps that serve as robust geometric conditioning for diffusion-based novel view synthesis. Our approach models depth in an angular domain using a localized Gaussian Process formulation, enabling computationally efficient inference while explicitly quantifying uncertainty in regions with limited observations. The reconstructed depth and uncertainty are used as a drop-in replacement for monocular depth estimators in existing diffusion-based rendering pipelines, without modifying the generative model itself. Experiments on real-world multimodal driving scenes demonstrate that replacing vision-only depth with our sparse range–based reconstruction substantially improves both geometric consistency and visual quality in single-image novel-view video generation. These results highlight the importance of reliable geometric priors for diffusion-based view synthesis and demonstrate the practical benefits of multimodal sensing even at extreme levels of sparsity. Code is publicly available at \href{https://github.com/importAmir/MultiModalNVS}{github.com/importAmir/MultiModalNVS}.
\end{abstract}

%% file: Sec/2_Intro.tex
\section{Introduction}
Synthesizing accurate images from novel viewpoints, commonly referred to as novel-view synthesis, is a fundamental problem with broad applications in virtual reality, robotics, and autonomous systems. Accurate geometric representations are critical for producing visually consistent novel views, particularly when only limited visual observations are available. While novel-view synthesis and 3D scene rendering have been studied for decades, reconstruction-based approaches such as Neural Radiance Fields (NeRFs) \citep{mildenhall2021nerf, barron2021mip} and Gaussian Splatting (GS) \citep{kerbl20233d, polyzos2025activeinitsplat, bao20253d} have recently demonstrated impressive rendering fidelity by explicitly modeling scene geometry from multi-view observations. However, these methods typically require dense image sets with high viewpoint coverage to achieve high-quality reconstructions, making them impractical in settings where only sparse or single-view observations are available.

To address sparse-view scenarios, generative approaches have emerged as an alternative to reconstruction-based models, aiming to synthesize plausible novel views without explicitly recovering full 3D scene representations. In the particularly challenging single-image setting, recent diffusion-based rendering pipelines \citep{liu2024reconx, yu2024viewcrafter, muller2024multidiff, ren2025gen3c} typically operate by first estimating depth from the input image and constructing an intermediate 3D representation, such as a point cloud, which is rendered along a target camera trajectory. A diffusion model is then conditioned on these rendered views to hallucinate missing content in disoccluded or unobserved regions, producing visually coherent novel views. While this paradigm has shown strong empirical performance compared to reconstruction-based methods in single-view settings, its effectiveness critically depends on the accuracy and consistency of the underlying depth estimates.

Monocular depth estimation from a single RGB image is inherently ill-posed, and existing approaches \citep{yang2024depth, wang2025moge} rely heavily on learned visual priors. In real-world environments, factors such as weak texture, challenging illumination, adverse weather, and occlusions frequently lead to depth predictions that are inaccurate or spatially inconsistent. In diffusion-based novel-view synthesis pipelines, these errors are not isolated: they are amplified through geometric back-projection and rendering, propagating across viewpoints and resulting in misalignment artifacts, inconsistent geometry, and degraded temporal coherence in the generated views. This observation highlights the need for improving the robustness of geometric initialization to enable reliable single-image novel-view synthesis.

In contrast to existing vision-only approaches, in the present work we introduce a multimodal diffusion-based approach for efficient novel view synthesis, whose contributions can be contextualized in the following aspects:
% We demonstrate that incorporating even extremely sparse multimodal range measurements provides a simple yet effective solution to this challenge. We propose a multimodal framework that leverages sparse radar or LiDAR measurements to reconstruct dense depth maps that serve as robust geometric priors for diffusion-based novel-view synthesis. Our approach formulates depth reconstruction in an angular domain using a localized Gaussian Process model, enabling computationally efficient inference while explicitly modeling uncertainty in regions with limited observations. Importantly, the proposed depth reconstruction acts as a drop-in replacement for monocular depth estimators and integrates seamlessly with existing diffusion-based rendering pipelines without modifying the generative model.

\begin{itemize}
    \item We introduce a range-sensor–based depth reconstruction module that leverages sparse radar or LiDAR measurements and serves as a drop-in replacement for vision-only monocular depth estimators in diffusion-based novel-view synthesis pipelines, while remaining agnostic to the diffusion model itself.
    
    \item Using only sparse range measurements, we propose an efficient depth reconstruction approach based on localized Gaussian Process modeling. By partitioning the image into spatially localized regions and fitting independent local Gaussian Processes, our method achieves improved computational efficiency while producing dense depth estimates with well-calibrated uncertainty.

    \item Experiments on real-world multimodal autonomous driving data demonstrate consistent improvements over image-only baselines. Replacing monocular depth with our multimodal reconstruction reduces LPIPS by 23.5\% and FID by 46.0\% in single-image novel-view video generation, while also improving depth accuracy, reducing mean absolute error by 4.5\% when evaluated against LiDAR ground truth.

\end{itemize}

%% file: Sec/4_Method.tex
\section{Method}
\subsection{Geometry-Conditioned Diffusion for Novel-View Synthesis}
Diffusion models generate samples by learning to reverse a fixed noising process.
Given a clean sample $x_0 \sim p_{\text{data}}(x)$, the forward diffusion process produces
\begin{equation}
x_\tau = \alpha_\tau x_0 + \sigma_\tau \epsilon, 
\qquad \epsilon \sim \mathcal{N}(0,I),
\end{equation}
where $\tau \in [0,1]$ denotes the diffusion time and $\{\alpha_\tau,\sigma_\tau\}$ follow a
predefined noise schedule.
A denoising network $f_\theta$ is trained to predict the injected noise, conditioned on auxiliary
information $c$, by minimizing
\begin{equation}
\mathcal{L}_{\text{diff}}(\theta)
=
\mathbb{E}_{x_0,\epsilon,\tau}
\left[
\left\|
f_\theta(x_\tau, \tau; c) - \epsilon
\right\|_2^2
\right].
\end{equation}
After training, novel samples are generated by iteratively applying the learned reverse process,
starting from Gaussian noise and guided by the conditioning signal. In this work, we use a standard
diffusion formulation and do not modify the generative model.

\begin{figure*}[t]
    \centering
    \includegraphics[width= \textwidth]{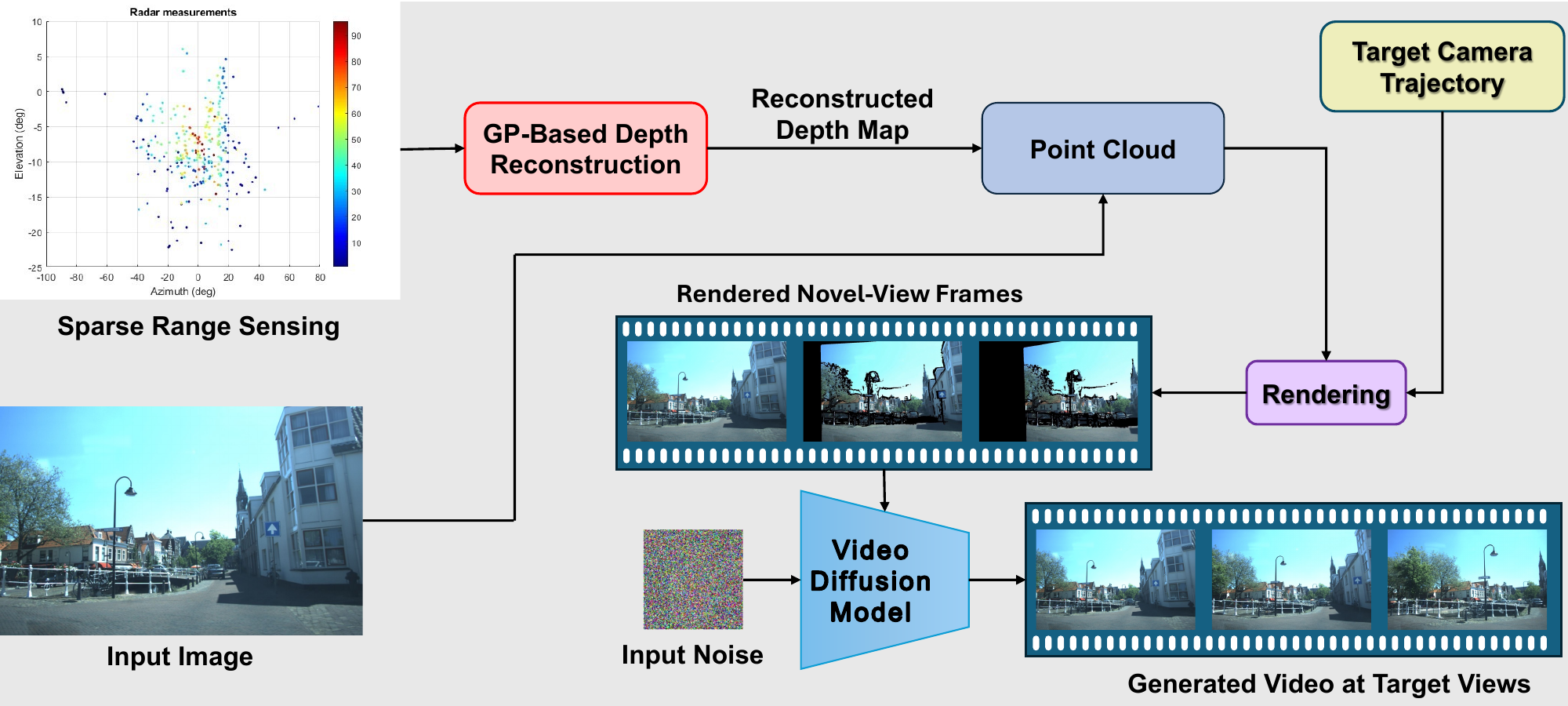}
    \vspace{-0.6cm}
    \caption{
    Overview of the proposed multimodal single-image novel-view synthesis pipeline. Sparse range sensing measurements are first processed by the proposed GP-based depth reconstruction module to produce a dense depth map. The reconstructed depth and the input RGB image are used to form a colored 3D point cloud, which is rendered along a target camera trajectory to generate sparse novel-view conditioning frames. These rendered frames are provided as geometric conditioning to a diffusion model, which synthesizes a temporally consistent video at the target viewpoints.
    }
    \vspace{-0.5cm}
    \label{fig:visual_results}
\end{figure*}

In sparse-view novel-view synthesis, the conditioning signal $c$ is derived from an explicit
geometric initialization in the form of rendered novel-view frames.
Given an input image $I \in \mathbb{R}^{H\times W \times 3}$ and camera intrinsics
$\mathbf{K}\in\mathbb{R}^{3\times 3}$, a depth estimator $D(\cdot)$ produces a depth map
$Z = D(I)$ for pixel coordinates $(u,v)$.
Let $\tilde{\mathbf{p}}=[u,v,1]^\top$ denote homogeneous image coordinates.
The corresponding 3D point in the camera frame is obtained via standard back-projection:
\begin{equation}
\mathbf{p}_{\text{cam}}(u,v)
=
\begin{bmatrix}
Z(u,v)\,\frac{u-c_x}{f_x}\\
Z(u,v)\,\frac{v-c_y}{f_y}\\
Z(u,v)
\end{bmatrix}.
\label{eq:backproj}
\end{equation}
Associating each $\mathbf{p}_{\text{cam}}(u,v)$ with its pixel color $I(u,v)$ yields a colored point
cloud
\begin{equation}
\mathcal{P}=\left\{\big(\mathbf{p}_{\text{cam}}(u,v),\, I(u,v)\big)\;:\;(u,v)\in\Omega\right\},
\end{equation}
where $\Omega$ denotes the set of valid pixels.
Given a target camera pose $\mathbf{T}_t = [\mathbf{R}_t \mid \mathbf{t}_t] \in SE(3)$, each 3D point
is transformed to the target camera frame:
\begin{equation}
\mathbf{p}^{(t)}_{\text{cam}}(u,v)
= \mathbf{R}_t\,\mathbf{p}_{\text{cam}}(u,v) + \mathbf{t}_t.
\label{eq:pose_transform}
\end{equation}
The transformed points are projected to the target image plane using the pinhole camera model:
\begin{equation}
\mathbf{p}^{(t)} =
\begin{bmatrix}
u^{(t)}\\ v^{(t)}
\end{bmatrix}
=
\begin{bmatrix}
f_x\,x^{(t)}/z^{(t)} + c_x\\
f_y\,y^{(t)}/z^{(t)} + c_y
\end{bmatrix}.
\label{eq:projection}
\end{equation}
Finally, the conditioning frame $c_t \in \mathbb{R}^{H\times W \times 3}$ is obtained by splatting
each colored point $(\mathbf{p}^{(t)}, I(u,v))$ onto the target image plane:
\begin{equation}
c_t = \mathrm{Splat}\!\left(\left\{\big(\mathbf{p}^{(t)}(u,v),\, I(u,v)\big)\right\}_{(u,v)\in\Omega}\right).
\label{eq:splat}
\end{equation}
For a target camera trajectory consisting of $T$ viewpoints, the diffusion model is conditioned on
the sequence of rendered frames $c = \{c_t\}_{t=1}^{T}$.
These conditioning frames depend entirely on the estimated depth $Z$.
Errors in $Z$ induce geometric misalignment across viewpoints, which propagate through the diffusion
process and result in view-dependent artifacts and degraded temporal consistency in the synthesized
views.
In this work, we replace the image-only depth estimator $D(\cdot)$ with a range-sensor–driven depth
reconstruction module, while keeping the unprojection, rendering, and diffusion components
unchanged.
The depth reconstruction is performed independently of the diffusion model and serves as a
drop-in geometric prior. An overview of the full geometry-conditioned diffusion pipeline is shown in Fig.~\ref{fig:visual_results}.

\subsection{Sparse Range-Sensor Depth Estimation via Gaussian Processes}
\label{sec:gp_depth}

\begin{figure*}[h]
    \centering
    \includegraphics[width= 0.9\textwidth]{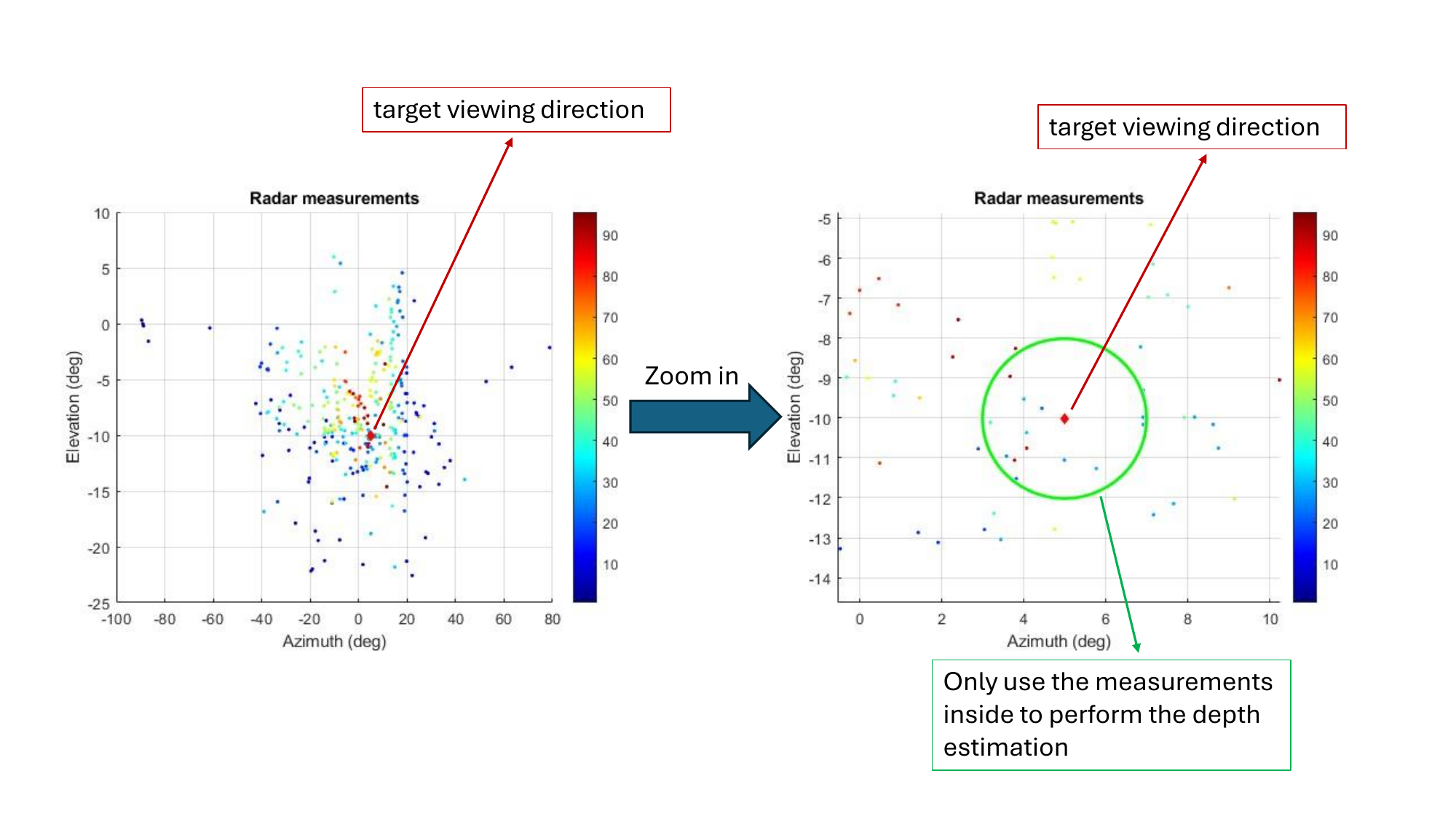}
    \vspace{-0.6cm}
    \caption{Illustration of the proposed localized Gaussian Process depth reconstruction in the angular domain. Left: Sparse radar range measurements represented in azimuth–elevation space, with the current target viewing direction indicated. Right: Zoomed-in view around the target location, highlighting the local neighborhood used for Gaussian Process inference. Only range measurements within this localized region contribute to the depth estimation at the target direction, enabling efficient and spatially adaptive depth reconstruction from sparse observations. 
    }
    \vspace{-0.5cm}
    \label{fig:localGP}
\end{figure*}

We consider a single acquisition from a sparse range sensing modality, such as radar or LiDAR,
which provides an unordered set of 3D range points without color information. These points are
mapped to azimuth and elevation angles, yielding a set of sparse depth measurements
$\mathbf{z}_T = [z_1,\ldots,z_T]$ at corresponding sensing directions.
Our objective is to reconstruct a dense depth map aligned with the image plane of a single RGB
input, while explicitly modeling uncertainty in regions with limited or no range observations.

To establish a geometrically consistent representation, we operate in an angular domain shared by
both sparse range measurements and image pixels.
Each range measurement is represented by its azimuth and elevation angles
$\mathbf{a}_t=(\phi_t,\theta_t)$ and its measured depth $z_t$.
For each image pixel $(u,v)$, we compute the corresponding camera ray using known intrinsics and
convert it to the same angular parameterization. The normalized camera ray is given by
\begin{equation}
\mathbf{r}(u,v) =
\begin{bmatrix}
(u - c_x)/f_x \\
(v - c_y)/f_y \\
1
\end{bmatrix}.
\end{equation}
The azimuth and elevation angles are then computed as
\begin{equation}
\phi(u,v) = \arctan\!\left(\frac{r_x}{r_z}\right), \qquad
\theta(u,v) = \arctan\!\left(\frac{r_y}{\sqrt{r_x^2 + r_z^2}}\right),
\end{equation}
where $\mathbf{r}(u,v) = [r_x, r_y, r_z]^\top$.
This representation naturally aligns sparse range observations with dense image pixels and avoids
projection ambiguities.

We model depth as a latent function $Z(\mathbf{a})$ defined over the angular domain and adopt
Gaussian Process (GP) regression \citep{rasmussen2006gaussian} to infer dense depth values from
sparse observations.
We place a GP prior on $Z(\mathbf{a})$ with a radial basis function (RBF) kernel and assume
independent Gaussian measurement noise:
\begin{equation}
Z \sim \mathcal{GP}(0,\kappa),
\qquad
z_t = Z(\mathbf{a}_t) + n_t,\ \ n_t \sim \mathcal{N}(0,\sigma_n^2).
\end{equation}
Given sparse observations $\{(\mathbf{a}_t, z_t)\}_{t=1}^T$, the posterior predictive distribution
at a query angular location $\mathbf{a}_\star$ is Gaussian,
\begin{equation}
p\!\left(Z(\mathbf{a}_\star)\mid \mathbf{A}_T,\mathbf{z}_T\right)
= \mathcal{N}\!\left(\mu_T(\mathbf{a}_\star),\,\sigma_T^2(\mathbf{a}_\star)\right),
\end{equation}
where $\mathbf{A}_T=[\mathbf{a}_1,\ldots,\mathbf{a}_T]$ and
$\mathbf{z}_T=[z_1,\ldots,z_T]^\top$.
The posterior mean and variance are
\begin{align}
\mu_T(\mathbf{a}_\star)
&= \mathbf{k}_T(\mathbf{a}_\star)^\top
\left(\mathbf{K}_T+\sigma_n^2\mathbf{I}\right)^{-1}\mathbf{z}_T, \\
\sigma_T^2(\mathbf{a}_\star)
&= \kappa(\mathbf{a}_\star,\mathbf{a}_\star)-
\mathbf{k}_T(\mathbf{a}_\star)^\top
\left(\mathbf{K}_T+\sigma_n^2\mathbf{I}\right)^{-1}\mathbf{k}_T(\mathbf{a}_\star),
\end{align}
where $[\mathbf{K}_T]_{m,m'}=\kappa(\mathbf{a}_m,\mathbf{a}_{m'})$ and
\(
\mathbf{k}_T(\mathbf{a}_\star)=
\big[\kappa(\mathbf{a}_1,\mathbf{a}_\star),\ldots,\kappa(\mathbf{a}_T,\mathbf{a}_\star)\big]^\top
\).

The posterior mean serves as the reconstructed depth, while the predictive variance quantifies
uncertainty. Computing the full GP posterior scales as $\mathcal{O}(T^3)$ in general. However,
depth is locally smooth in the angular domain, and depth at a given viewing direction is primarily
influenced by nearby range measurements, motivating a localized formulation.

We therefore adopt a per-query localized Gaussian Process formulation.
For each query angular location $\mathbf{a}_\star$, we define a local neighborhood
\begin{equation}
\mathcal{R}(\mathbf{a}_\star)
=
\left\{
\mathbf{a}_t : \|\mathbf{a}_t - \mathbf{a}_\star\|_2 \le r
\right\},
\end{equation}
where $r$ is a fixed angular radius.
Only range measurements within this circular neighborhood contribute to GP inference at
$\mathbf{a}_\star$.
Let $\mathbf{A}_{T_\star}$ and $\mathbf{z}_{T_\star}$ denote the measurements within
$\mathcal{R}(\mathbf{a}_\star)$, with $T_\star \ll T$ due to sparse sensing.

For each query, we use an RBF kernel
\begin{equation}
\kappa(\mathbf{a},\mathbf{a}')
= \sigma_f^2 \exp\!\left(-\frac{1}{2\ell^2}
\|\mathbf{a}-\mathbf{a}'\|_2^2\right),
\end{equation}
where $\ell$ controls angular smoothness.
The signal variance $\sigma_f^2$ is fixed based on prior knowledge of the sensing range, while the
length scale $\ell$ is selected via marginal likelihood maximization.
For each image pixel $(u,v)$, we query the local GP centered at $\mathbf{a}(u,v)$ to obtain a dense
depth estimate and its associated uncertainty. This per-query localized formulation reduces
computational complexity to $\mathcal{O}(T_\star^3)$ per query and is trivially parallelizable
across image locations. Fig.~\ref{fig:localGP} illustrates the localized GP formulation.

The predictive variance $\sigma_{T_\star}^2(\mathbf{a}_\star)$ provides a principled measure of
depth reliability.
During geometric rendering, depth estimates whose variance exceeds a fixed threshold are masked
out, preventing unreliable geometry from contributing to the conditioning frames used by the
diffusion model.
This uncertainty-aware depth reconstruction yields more stable geometric conditioning and improves
the robustness and consistency of downstream novel-view synthesis.

\noindent \textbf{Remark 1}. In this work, we consider the RBF kernel for GP-based modeling. While existing works aim to identify the most appropriate kernel function $\kappa$ to effectively capture the covariance between function evaluations; see e.g., \cite{lu2023surrogate}, they could be readily utilized in our proposed approach but this exceeds the scope of the current work. 

\noindent \textbf{Remark 2}. Although conventional GPs incur $\mathcal{O}(T^3)$ complexity that grows rapidly with the number of measurements $T$, the proposed localized GP framework reduces the cost to $\mathcal{O}(T_\star^3)$ per neighborhood, where $T_\star \ll T$. Moreover, all neighborhood GPs can be processed in parallel, further improving computational efficiency.

% We therefore partition the angular domain into $M$ non-overlapping regions
% $\{\mathcal{R}_1,\ldots,\mathcal{R}_M\}$ and fit an independent local GP within each region.
% For region $\mathcal{R}_r$, let $\mathbf{A}_{T_r}$ and $\mathbf{z}_{T_r}$ denote the measurements
% falling inside the region, with $T_r \ll T$.
% We adopt a region-specific RBF kernel
% \begin{equation}
% \kappa_r(\mathbf{a},\mathbf{a}')
% = \sigma_f^2 \exp\!\left(-\frac{1}{2\ell_r^2}
% \|\mathbf{a}-\mathbf{a}'\|_2^2\right),
% \end{equation}
% where $\ell_r$ controls angular smoothness.
% The signal variance $\sigma_f^2$ is fixed based on prior knowledge of the sensing range, while the
% length scale $\ell_r$ is estimated by maximizing the regional log marginal likelihood. For each image pixel $(u,v)$, we query the GP corresponding to the region containing
% $\mathbf{a}(u,v)$ to obtain a dense depth estimate and its associated uncertainty.
% This localized formulation reduces computational complexity to $\mathcal{O}(T_r^3)$ per region and
% enables parallel inference across regions. 

%% file: Sec/5_Experiments.tex
\section{Experiments}
\subsection{Dataset}
We evaluate our method using the View-of-Delft (VoD) dataset \citep{palffy2022multi}, a multi-modal autonomous driving dataset containing synchronized automotive radar, camera, and LiDAR data collected in urban environments. For evaluation, we curate a subset of 26 diverse video segments
spanning a range of urban scene types and capture conditions, which we use to benchmark single-image
novel-view synthesis performance.

\subsection{Single View to Video Generation}
We use the camera metadata provided with each VoD sequence to define a target camera trajectory for
novel-view synthesis. Concretely, we take the first frame of each sequence as the input reference
image and use the recorded camera poses from this first frame through the final frame to specify
the target trajectory along which novel views are synthesized. For geometry estimation, our method
reconstructs depth from synchronized sparse range measurements using the proposed Gaussian
process–based depth reconstruction module. We evaluate two multimodal variants: one using sparse
automotive radar measurements, which correspond to approximately 0.02\% of image pixels, and one
using sparse LiDAR measurements, which correspond to approximately 0.52\% of image pixels. In
contrast, the image-only baseline follows the default GEN3C pipeline \citep{ren2025gen3c} and
infers depth from the RGB input using the monocular depth estimator MoGe \citep{wang2025moge}.
In all cases, the estimated depth is back-projected to form a colored 3D point cloud, which is
rendered along the target camera trajectory to produce sparse novel views with disoccluded and
unobserved regions. A diffusion-based inpainting stage then completes missing content to generate
a temporally consistent video along the specified camera path.

\begin{figure*}[!t]
    \centering
    \includegraphics[width= 0.9\textwidth]{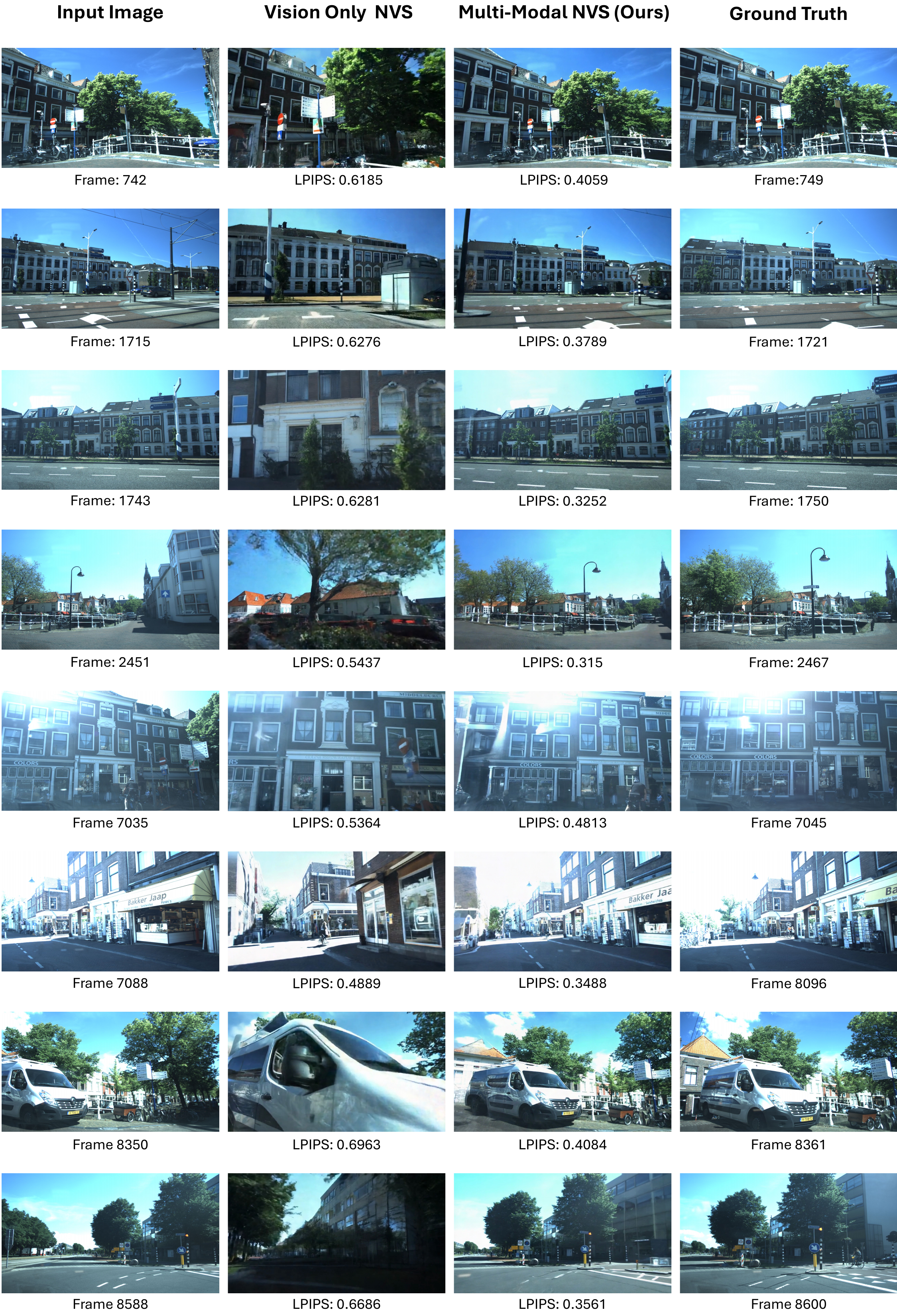}
    \caption{
    Qualitative comparison on single-image novel-view synthesis on the View-of-Delft dataset.
    From left to right, we show the input image, the novel view generated by GEN3C \citep{ren2025gen3c} using its default
    monocular depth estimator, MoGe \citep{wang2025moge}, the novel view generated by GEN3C when replacing monocular depth with our
    sparse range-sensor depth reconstruction module, and the ground-truth target view.
    For each generated view, we report LPIPS with respect to the ground truth (lower is better).
    Across all examples, our depth reconstruction yields consistently lower LPIPS and improved geometric alignment, underscoring the importance of reliable geometry for diffusion-based rendering from single-view inputs.
    }
    %\vspace{-0.5cm}
    \label{fig:visual_results}
\end{figure*}

\begin{table}[t]
    \captionsetup{justification=raggedright,singlelinecheck=false}
    \caption{
Quantitative evaluation on the View-of-Delft dataset for single-image novel-view video generation
with GEN3C \citep{ren2025gen3c} under three depth-conditioning variants:
(i) the default pipeline using the vision-only monocular depth estimator MoGe \citep{wang2025moge},
(ii) the same pipeline with monocular depth replaced by our multimodal sparse range-sensor depth
reconstruction module using synchronized radar measurements (approximately 0.02\% pixel coverage),
and (iii) the same replacement using synchronized LiDAR measurements (approximately 0.52\% pixel coverage).
Metrics are computed with respect to ground-truth target views, and the results show consistent improvements across
all measures when using our reconstructed depth, emphasizing the role of reliable depth priors in
diffusion-based novel-view synthesis from a single image.
}
    \label{tab:depth_estimator_comparison}
    \vspace{2pt}
    \resizebox{\linewidth}{!}{%
    \begin{tabular}{lcccccc}
        \toprule
        Method &
        Pixel coverage &
        $\mathrm{PSNR}\,\uparrow$ &
        $\mathrm{SSIM}\,\uparrow$ &
        $\mathrm{LPIPS}\,\downarrow$ &
        $\mathrm{FID}\,\downarrow$ &
        $\mathrm{t-LPIPS}\,\downarrow$ \\
        \midrule
        GEN3C (Vision-Only Monocular Depth) & — & 12.36 & 0.4561 & 0.5804 & 152.62 & 0.1117 \\
        % Depth Anything V2 & 15.03 & 0.4950 & 0.4182\\
        GEN3C (Ours, Multi-Modal with radar) & 0.02\% & 14.26 & 0.4860 & 0.4441 & 82.41 & 0.0790 \\
        GEN3C (Ours, Multi-Modal with LiDAR) & 0.52\% & \textbf{14.69} & \textbf{0.4971} & \textbf{0.4230} & \textbf{71.91} & \textbf{0.0563} \\
        \bottomrule
    \end{tabular}%
   }
\end{table}
For evaluation, we compare the generated videos against the corresponding ground-truth frames
captured along the same trajectory. We report pixel-aligned PSNR and SSIM, perceptual similarity
via LPIPS~\citep{zhang2018unreasonable}, distributional quality via FID~\citep{heusel2017gans}, and
temporal consistency via temporal LPIPS. Quantitative results are summarized in
Table~\ref{tab:depth_estimator_comparison}. Replacing monocular depth with our multimodal depth
reconstruction consistently improves performance across all metrics. Using sparse radar
measurements, PSNR increases from 12.36 to 14.26 ($\approx$15.4\%), SSIM increases from 0.4561 to
0.4860 ($\approx$6.6\%), LPIPS decreases from 0.5804 to 0.4441 ($\approx$23.5\% reduction), FID
decreases from 152.62 to 82.41 ($\approx$46.0\% reduction), and temporal LPIPS decreases from
0.1117 to 0.0790 ($\approx$29.3\% reduction). Using sparse LiDAR measurements yields further
improvements, achieving a PSNR of 14.69, SSIM of 0.4971, LPIPS of 0.4230, FID of 71.91, and
temporal LPIPS of 0.0563. These results indicate that improving the reliability of geometric priors, particularly through
even sparse range sensing—translates directly into higher-fidelity and more temporally consistent
diffusion-based novel-view synthesis from a single image. We further provide qualitative
comparisons in Fig.~\ref{fig:visual_results}, which corroborate these trends by showing improved
geometric alignment and reduced view-dependent artifacts when using our depth reconstruction.

\subsection{Depth Estimation Accuracy}
\begin{table}[t]
    \captionsetup{justification=raggedright,singlelinecheck=false}
    \caption{
    Depth estimation accuracy on the View-of-Delft dataset evaluated on the first frame of each of the
    26 selected segments. Our multimodal depth reconstruction uses synchronized radar measurements that
    cover approximately 0.02\% of image pixels. Predicted depth is compared against LiDAR-derived depth
    at pixels where LiDAR measurements are available. Lower is better for both metrics.
    }
    \label{tab:depth_estimator_depth_metrics}
    \centering
    \vspace{-0.2cm}
    \resizebox{0.6\linewidth}{!}{%
    \begin{tabular}{lcc}
        \toprule
        Method &
        $\mathrm{MAE}\,\downarrow$ &
        $\mathrm{RMSE}_{\log}\,\downarrow$ \\
        \midrule
        MoGe 
        & 14.25 & 2.23 \\
        Depth Anything V2 & 18.70 & 0.94 \\
        Ours (Sparse Radar Depth) & \textbf{13.61} & \textbf{0.92} \\
        \bottomrule
    \end{tabular}%
    }
\end{table}

To directly assess the quality of the reconstructed depth, we evaluate depth estimation accuracy
on the View-of-Delft dataset using LiDAR-derived depth as ground truth. We conduct this evaluation on the first image of each of the 26 selected scenarios. 
Depth predictions are evaluated at image pixels where LiDAR measurements are available, ensuring a
fair and consistent comparison across methods.

We compare our sparse radar–based depth reconstruction against two representative monocular depth
estimators: MoGe \citep{wang2025moge} and Depth Anything V2 \citep{yang2024depth}.
All methods produce a dense depth map aligned with the reference image, which is then compared to
the corresponding LiDAR depth values at valid pixels.
We report mean absolute error (MAE) in linear depth and root mean squared error in log depth
(RMSE$_{\log}$), with lower values indicating more accurate depth estimation.

Quantitative results are summarized in Table~\ref{tab:depth_estimator_depth_metrics}.
Our method achieves the lowest error across both metrics, improving MAE from 14.25 to 13.61
($\approx$4.5\% relative reduction over the best monocular baseline, MoGe) and improving
$\mathrm{RMSE}_{\log}$ from 0.94 to 0.92 ($\approx$2.1\% relative reduction over the best monocular
baseline, Depth Anything V2). These results indicate that incorporating sparse radar measurements
yields more accurate and reliable depth estimates than vision-only monocular approaches.
\subsection{Ablation Studies}
\subsubsection{Impact of Uncertainty-Based Masking on Novel-View Synthesis}
We further evaluate the effect of uncertainty-based masking on downstream novel-view synthesis. For this experiment, we use radar transmissions as input and conduct the study on 9 videos from the VoD dataset. We then vary the fraction of pixels retained according to the predictive uncertainty produced by our depth reconstruction module. Specifically, after estimating a dense depth map together with its per-pixel predictive variance, we rank pixels by uncertainty and retain only the $p\%$ most certain pixels, while masking out the remaining high-uncertainty regions. This experiment is designed to isolate the effect of uncertainty filtering on the quality of the conditioning signal provided to the diffusion model.

The results are reported in Table~\ref{tab:uncertanity_masking}. We observe that uncertainty masking yields a clear trade-off between geometric reliability and conditioning coverage. When the retained percentage is too low, the conditioning frames are composed of highly reliable geometry, but they become excessively sparse and provide insufficient structural information to the diffusion model. As a result, the model must hallucinate a larger portion of the scene, which limits its ability to preserve accurate scene layout and appearance across viewpoints. At the other extreme, retaining all pixels maximizes coverage but also reintroduces unreliable depth estimates in regions of high uncertainty. These noisy predictions lead to inconsistent geometric cues during rendering, which in turn degrade perceptual quality and temporal coherence.

\begin{table}[h]
    \centering
    \captionsetup{justification=raggedright,singlelinecheck=false}
    \caption{Ablation on uncertainty-based masking for depth-conditioned novel-view synthesis. After reconstructing dense depth and predictive uncertainty, we retain only the $p\%$ most certain pixels and mask out the remaining high-uncertainty regions before rendering the conditioning frames. Results on 9 videos from the View-of-Delft dataset show that retaining $80\%$ of the lowest-uncertainty pixels provides the best overall trade-off between geometric reliability and conditioning coverage, leading to the strongest downstream generation performance.}
    \label{tab:uncertanity_masking}
    \vspace{2pt}
    \resizebox{0.7\linewidth}{!}{
    \begin{tabular}{lccccc}
        \toprule
        Retention Ratio &
        $\mathrm{PSNR}\,\uparrow$ &
        $\mathrm{SSIM}\,\uparrow$ &
        $\mathrm{LPIPS}\,\downarrow$ &
        $\mathrm{FID}\,\downarrow$ &
        $\mathrm{t-LPIPS}\,\downarrow$ \\
        \midrule
        $40\% $ & $14.36$ & $0.4713$ & $0.4143$ & $78.46$ & $0.0583$ \\
        $60\%$ & $14.72$ & $0.4747$ & $0.4136$ & $69.82$ & $0.0583$ \\
        $80\%$ & $\textbf{14.80}$ & $\textbf{0.4826}$ & $\textbf{0.4029}$ & $\textbf{67.91}$ & $\textbf{0.0578}$ \\
        $100\%$ & $14.70$ & $0.4800$ & $0.4139$ & $78.44$ & $0.0648$\\
        \bottomrule
    \end{tabular}
   }
\end{table}
Among all settings, retaining $80\%$ of the lowest-uncertainty pixels provides the most favorable overall trade-off, indicating that it preserves sufficient geometric coverage for effective diffusion conditioning while filtering out the most unreliable depth estimates. Overall, these results suggest that moderate uncertainty filtering improves the fidelity and consistency of the synthesized views by balancing confidence and coverage in the reconstructed geometry. Based on this ablation, we retain the $80\%$ most certain pixels in all experiments.

\subsubsection{Ablation on Locality Radius}

We further study the effect of the locality radius used in our depth estimation module. For this ablation, we use the same 9 videos from the View-of-Delft dataset as in the previous study. We vary the angular locality radius \(r \in \{1^\circ, 2^\circ, 4^\circ\}\) when estimating dense depth from sparse radar observations, and evaluate the resulting predictions against LiDAR depth on valid pixels using the same protocol as in Table~\ref{tab:depth_estimator_depth_metrics}. Here, the locality radius defines the size of the local neighborhood in angular space, parameterized by azimuth and elevation. For each target pixel, we first convert its camera ray into an azimuth--elevation representation, and then gather sparse radar points whose angular distance to that ray falls within radius \(r\). The depth at that pixel is then inferred from this local set of neighboring observations. In this way, a smaller radius enforces stronger locality but uses fewer supporting points, whereas a larger radius incorporates more observations at the expense of reduced locality. We report MAE and \(\mathrm{RMSE}_{\log}\) to analyze how the size of this angular neighborhood affects depth accuracy.

The results are summarized in Table~\ref{tab:abilation_radius}. We observe that a locality radius of \(2^\circ\) achieves the best overall performance, yielding the lowest MAE and \(\mathrm{RMSE}_{\log}\). When the radius is reduced to \(1^\circ\), the prediction relies on a more limited local neighborhood, which restricts the amount of geometric evidence available for inference and leads to slightly degraded accuracy. In contrast, increasing the radius to \(4^\circ\) does not improve performance and instead slightly worsens both metrics, while also increasing computational cost since more sparse points fall within each local neighborhood and must be processed during Gaussian Process inference. Therefore, \(2^\circ\) provides the best trade-off between reconstruction accuracy and efficiency, and we use \(2^\circ\) in all experiments.

\begin{table}[h]
\captionsetup{justification=raggedright,singlelinecheck=false}
    \caption{Ablation on the angular locality radius used in dense depth estimation. Using the same 9 videos from the View-of-Delft dataset as in the previous ablation, we vary the local neighborhood radius \(r\) when predicting dense depth from sparse radar observations and evaluate the predictions against LiDAR ground truth on valid pixels. A radius of \(2^\circ\) achieves the best overall performance, while increasing the radius to \(4^\circ\) does not improve accuracy and incurs higher computational cost due to the larger number of local points considered during Gaussian Process inference.}
    \label{tab:abilation_radius}
    \centering
    \resizebox{0.4\linewidth}{!}{%
    \begin{tabular}{lcc}
        \toprule
        Locality Radius &
        $\mathrm{MAE}\,\downarrow$ &
        $\mathrm{RMSE}_{\log}\,\downarrow$ \\
        \midrule
        r = 1 & 10.88 & 0.646 \\
        r = 2 & \textbf{10.67} & \textbf{0.627} \\
        r = 4 & 10.94 & 0.648 \\
        \bottomrule
    \end{tabular}%
    }
\end{table}